\documentclass[runningheads]{llncs}

\usepackage[T1]{fontenc}
\usepackage{graphicx}
\usepackage{url}
\usepackage{booktabs}
\usepackage{xcolor}
\usepackage{algorithm}
\usepackage{algpseudocode}
\usepackage{amsmath,amssymb}
\usepackage{enumitem}
\usepackage[hidelinks]{hyperref}

\newcommand{\fld}[1]{\textsf{#1}}
\usepackage{tabularx}
\usepackage{booktabs}

\usepackage{colortbl}
\definecolor{oursrow}{HTML}{F4ECEA}
\usepackage{listings}
\lstdefinestyle{prompt}{%
  basicstyle=\ttfamily\footnotesize, breaklines=true, breakatwhitespace=true,
  columns=fullflexible, keepspaces=true, showstringspaces=false,
  frame=single, framesep=5pt, xleftmargin=6pt, xrightmargin=6pt,
}


\begin{document}

\title{PaperRouter-Agent: A Content-Grounded LLM Agent for Personalized Hierarchical Paper Routing}
\titlerunning{PaperRouter-Agent}

\author{Keshen~Zhou \and Lintao~Wang \and Suqin~Yuan \and Zhuqiang~Lu \and Yu~Luo \and Zhiyong~Wang}
\authorrunning{K. Zhou et al.}
\institute{University of Sydney, Sydney, NSW 2006, Australia\\
\email{\{kzho6770,lwan3720,zhlu6105,yluo0465\}@uni.sydney.edu.au}\\
\email{suqinyuan.cs@gmail.com, zhiyong.wang@sydney.edu.au}}

\maketitle

\begin{abstract}
Researchers organize the papers they collect into personal folder hierarchies in reference managers, and route each new paper into the folder where it belongs.
This task differs from standard hierarchical text classification. 
A user’s folder hierarchy is not a fixed, shared taxonomy but a private and evolving folksonomy whose folder meanings may be topical, shorthand, venue-based, or process-oriented, and are often defined by the papers already stored inside them.
We formalize this setting as \emph{personalized hierarchical paper routing} (PHPR): assigning an incoming paper to folders in a user-specific hierarchy without per-user training. 
We propose \textbf{PaperRouter-Agent}, a training-free LLM agent that grounds routing decisions in folder members rather than folder names alone. 
The agent first narrows the candidate hierarchy, retrieves folder-specific evidence, verifies fit by inspecting member papers, and incorporates similarity-gated feedback from past user rejections.
A formative study on real personal libraries shows that PaperRouter-Agent raises overall Recall@1 from $0.39$ to $0.61$ and Recall@3 from $0.57$ to $0.83$, with the largest gains on organizational folders defined by metadata such as venue or year, where single-shot methods collapses (Recall@1 $0.09\to0.50$).
On the public LaMP-2 benchmark, the same approach improves accuracy from $44.5\%$ to $51.5\%$ ($+9.0$ macro-F1) over a single-shot baseline, while remaining low-cost for practical use.
\keywords{Personalized text classification \and Retrieval-augmented generation \and LLM agents \and Personal knowledge management}
\end{abstract}
\section{Introduction}
\label{sec:intro}

\begin{figure}[t]
  \centering
  \includegraphics[width=\textwidth]{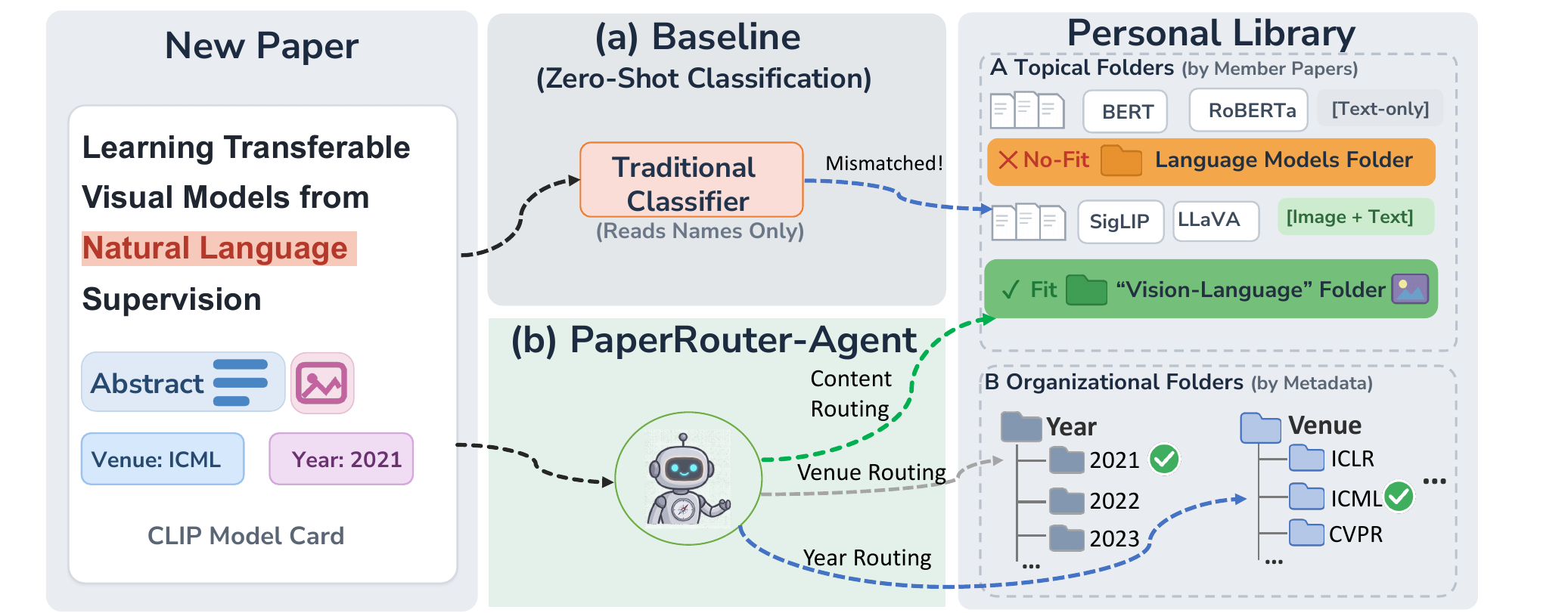}
  \caption{Example: Comparison of a name-based baseline and the proposed content-grounded PaperRouter-Agent. While the baseline misclassifies a multimodal paper based solely on its title, the proposed agent accurately routes it to appropriate topical and metadata-defined folders.}
  \label{fig:running-example}
\end{figure}

Organizing papers collected from the Web has become a routine burden of research practice. 
ArXiv's monthly submissions have grown from 24k in October 2024 to a recent peak of 30k in March 2026~\cite{arxiv_stats}, and surveys of reading behavior find that researchers read on the order of 20 scholarly articles per month~\cite{tenopir2019scholarlyreading}.
The papers worth keeping are added to personal reference managers such as Zotero and EndNote, where a library grows over years into a hierarchy of folders containing hundreds of papers.
New papers keep arriving and the hierarchy itself keeps changing, so routing each paper into the right folder is not a one-time effort but a recurring task that requires recognizing which existing folder already holds similar work.
Reference managers automate storage, search, and citation, but not this routing: which folder a paper belongs to is still determined by the user.
Automating this step is a personalized classification problem: the labels are each researcher's own folders, not a single schema shared across users. 

The difficulty of routing is that a folder's meaning is defined by the papers a user has placed in it, whereas its name is often the only signal available for an incoming paper.
In a personal library, folder names rarely follow a single organizing principle: some are topical (``Diffusion Models in Medical Imaging''), some are personal shorthand (``3D'', ``MoE''), and some record a principle other than topic, such as a venue or a reading status (``CVPR'', ``To Read'').
A zero-shot baseline that relies on names alone misclassifies CLIP~\cite{clip} because of this gap, as Fig.~\ref{fig:running-example} illustrates.
Standard hierarchical text classification (HTC) assumes a fixed \emph{taxonomy} shared across users, in which a name fully determines a label's meaning.
A personal library is instead an evolving \emph{folksonomy}~\cite{vanderwal2007folksonomy,golder2006usage} that each researcher defines and updates for their own use.
Deciding where an incoming paper belongs depends not only on a folder's often ambiguous name, but also on its contents.

To our knowledge, no prior work directly addresses routing into a personal, evolving hierarchy, although two related lines each cover part of it (Section~\ref{sec:related}).
HTC~\cite{zangari2024htc} reads a label from its name or an external description, never from its members.
Personalized LLM agents~\cite{zhang2024personalization,li2024personalagents} offer architectures for user profiles, memory, and planning, but address question answering and recommendation rather than routing into a user's own folders.
What neither provides is what this task requires: a folder interpreted by its members as well as its name, evidence selected by folder type, and a meaning learned from a user's own decisions when there is nothing to train on.
These requirements call for a system that proceeds in several steps and maintains state between them, beyond the reach of a single fixed classifier.
We formulate this setting as \textbf{personalized hierarchical paper routing (PHPR)}: assigning an incoming paper to one or more folders of a user's personal, evolving hierarchy, with no training on that user's library.
We address it with \textbf{PaperRouter-Agent}, a training-free LLM agent that routes by reading what each folder holds rather than its name.

PaperRouter-Agent carries out this routing in four stages.
Since a personal hierarchy can contain many folders, candidate narrowing (\textbf{Planner}) traverses it from the top and keeps a small set of candidates.
For each candidate, evidence retrieval (\textbf{Retriever}) selects evidence by type, content for a topical folder and metadata for a venue or status one.
Content-grounded verification (\textbf{Inspector}) then reads the members of each candidate folder and judges whether the incoming paper belongs with them.
Feedback-based reflection (\textbf{Reflector}) maintains a memory of which suggestions a user accepts or rejects and brings the relevant cases into later verification.
Throughout, the hierarchy is treated as the researcher's own organizing structure, to be understood rather than rebuilt.
Because routing into a personal hierarchy is a new and inherently personalized setting, we evaluate it from two complementary angles.
A study on researchers' own libraries surfaces where name-based routing fails and how users experience content-grounded routing.
An offline comparison on the public LaMP-2 benchmark then tests content-grounded inspection against a single-shot baseline.

In summary, the key contributions of our work are as follows: 
\begin{enumerate}

  \item \textbf{From taxonomy to folksonomy.} We formalize our task as  \emph{personalized hierarchical paper routing} (PHPR) and characterize the folder types and name-based failures that separate a personal \emph{folksonomy} from the shared \emph{taxonomy} of HTC.

  \item \textbf{A four-stage agent framework.} We propose PaperRouter-Agent, a four-stage LLM agent (Planner, Retriever, Inspector, Reflector) that solves PHPR through \emph{content-grounded inspection} and \emph{feedback-based reflection}.

  \item \textbf{Evaluation and results.} We evaluate PaperRouter-Agent on researchers' own libraries through formative studies and on the public LaMP-2 benchmark, where it achieves state-of-the-art results among training-free, retriever-free methods and matches a retrieval-augmented LLM.
  
\end{enumerate}

\section{Related Work}
\label{sec:related}
\subsection{Hierarchical Text Classification}

Hierarchical text classification (HTC) assigns each document to one or more nodes of a predefined label hierarchy~\cite{zangari2024htc}. 
The hierarchy is an expert-curated taxonomy shared across users, such as Web of Science~\cite{kowsari2017hdltex}, DBpedia~\cite{lehmann2015dbpedia}, or MeSH~\cite{lipscomb2000mesh}, and methods need less and less labeled data, from full supervision~\cite{zhou2020hiagm} to class-name-only supervision~\cite{shen2021taxoclass} to zero-shot prompting~\cite{schmidt2024singlepass}.
Recent LLM-based methods keep this fixed, shared taxonomy and change only how a document is classified into it: 
turning class names into prototypes~\cite{zhang2025hierprompt}, attaching knowledge-graph context~\cite{zhang2025teleclass}, or traversing the hierarchy step by step~\cite{yoshimura2025tmh}. 
Some methods revise the taxonomy itself, rewriting~\cite{golde2026taxmorph} or rebuilding~\cite{kargupta2025taxoadapt,li2026evotaxo} it to align with the large language model's knowledge. 
But they still operate on a single expert-curated taxonomy shared across all users, never a personalized one.
Models are also trained and evaluated offline on this shared corpus, not on a user's small and changing hierarchy.

A hierarchy that a researcher keeps fits these assumptions poorly (Table~\ref{tab:phpr-vs-htc}).
It is a folksonomy: user-defined, evolving, and built from mixed organizing principles such as topic, venue, and reading status, not a shared taxonomy. 
A folder named ``3D'' is defined by papers already in it, not only by its name, so routing a paper here means asking whether it belongs with those members. 
Traditional HTC classifiers are unlikely to address this challenge, as they work from class names, definitions, or prototypes rather than from the members themselves.
Even prototype methods build their summaries once and offline, whereas a personal library offers no training pass, its folders are few and constantly changing, and some (a venue or a reading status) have no topical prototype at all. 

PaperRouter-Agent instead decides each candidate folder from the folder's own evidence: it reads the member papers of a topical folder, and matches the incoming paper's metadata against a venue or year folder. 
In both cases, the decision depends on what the folder contains rather than its name alone.

\subsection{Personalization and LLM Agents}

Routing into a personal hierarchy needs an agent with two abilities that existing systems rarely combine.
It must decide for itself how to interpret each folder, since folders are heterogeneous, and it must keep learning from the user, whose only signal is accepting or rejecting each suggestion.
Per-user classification is the obvious comparison, but it provides neither: 
EmFore~\cite{singh2023emfore}, BYOC~\cite{bohra2023byoc}, Ding et al.~\cite{ding2022opentopic}, and end-user classifier studies~\cite{wang2025enduser} personalize the label set, yet they remain flat, with a single level rather than a hierarchy.
They judge each folder from its name or description, without understanding what it actually is, whether a topic, a venue, or a reading status.

Recent work on LLM agents already provides the necessary components.
Agentic frameworks structure multi-step reasoning~\cite{yao2023react,shinn2023reflexion}, ground each step in retrieved evidence~\cite{asai2024selfrag}, and adapt without retraining~\cite{memento2025}.
But they have not been put together for routing into a personal, evolving hierarchy: 
the nearest benchmark, LaMP-2~\cite{salemi2024lamp}, scores only offline personalized tagging, and multi-agent systems route among agent roles or reasoning steps, not a user's folders~\cite{agentrec2025}.
Reflective classifiers also rely on human feedback, but differ from our work in how that feedback is used, not in whether a human provides it.
Prior reflective classifiers refine themselves offline, through self-critique or a round of human annotation before deployment~\cite{hassell2025reflective}.
Our Reflector instead stores the user's accept and reject decisions in memory during use, and reuses the most similar past rejections when routing the next paper~\cite{sanzguerrero2025cicl}.

In practice, the closest systems are deployed tools.
Assistants that draw on a user's own papers~\cite{chang2023citesee,lee2024paperweaver} and recent Zotero AI tools~\cite{zoterollmtagger,papersgpt} tag items, answer questions, or reorganize a library.
But they hardly treat content-grounded routing into a personal hierarchy as a defined task or report an evaluation. 
The two abilities are also coupled: feedback has nothing to refine unless the system already reads folder contents.
A flat, name-based classifier therefore cannot simply add the missing piece.
In contrast, PaperRouter-Agent does both: it reads each folder's own contents to route, and it learns from the user's feedback online.

\section{Problem Statement and Motivation}
\label{sec:problem}

We call the task \emph{personalized hierarchical paper routing} (PHPR). 
Given

\begin{itemize}[leftmargin=*,itemsep=1pt,topsep=2pt]
  \item an incoming paper $p = (\text{title}, \text{abstract}, \text{metadata})$;
  \item a user's collection hierarchy $H_u$, each folder $f$ carrying $\langle\,$name, parent, \mbox{children}, \mbox{members}$\,\rangle$, where \emph{children} are subfolders and \emph{members} are the papers already filed in $f$;
  \item the user's past feedback $F_u = \{(p_i, f_j, \text{accept}\mid\text{reject}\mid\text{override})\}$,
\end{itemize}
\noindent Return a ranked set of folders $R \subseteq \mathrm{leaves}(H_u)$ for $p$. 

\noindent Routing is multi-target and ranked, and we report $\text{Recall@}K$ against the folders the user accepts. 
These labels are the user's own folksonomy, not a fixed taxonomy shared across users (Table~\ref{tab:phpr-vs-htc}).

\begin{table}[!htbp]
  \centering
  \caption{How Personalized Hierarchical Paper Routing (PHPR) differs from classical hierarchical text classification (HTC).}
  \label{tab:phpr-vs-htc}
  \footnotesize
  \setlength{\tabcolsep}{6pt}
  \renewcommand{\arraystretch}{1.15}
  \newcolumntype{Y}{>{\raggedright\arraybackslash}X}
  \begin{tabularx}{\linewidth}{@{}>{\bfseries}l Y >{\columncolor{oursrow}}Y@{}}
    \toprule
     & \textbf{HTC} & \textbf{PHPR (ours)} \\
    \midrule
    Label set            & Fixed, expert-curated, shared by all users   & User-defined, per-user, always changing \\
    Defined by           & The label's name or a description of it       & The folder's name \emph{and} the papers already in it \\
    Organizing principle & Always the topic                             & Topic, venue, or reading status \\
    Supervision          & Offline expert labels                        & The user's own accept/reject feedback, online \\
    \bottomrule
  \end{tabularx}
\end{table}

\textbf{Motivation.}
In a personal library, what a folder means is set by the papers inside it, while a new paper usually arrives with only its name.
A name that states its subject, such as \emph{Diffusion Models in Medical Imaging}, routes the paper on its own.
A shorthand name such as \emph{3D} or \emph{UKG} does not, and its subject can be inferred only from its member papers.
A venue or year, such as CVPR or 2024, is resolved by the paper's metadata rather than its content.
A reading-status label such as \emph{To Read} records what the user plans to do with a paper, a cue the paper itself does not carry.
We refer to these four cases as semantic-clear, semantic-shorthand, venue/year, and process/status, and report per-type routing accuracy in Section~\ref{sec:eval:user-study}.
Routing must therefore read what each folder holds and choose the evidence its type requires, rather than simply match names.

But reading folder contents is not a single classification step.
A personal hierarchy is deep and keeps changing, so the system first narrows the many folders to a small set of candidates.
Among these, it identifies what each folder is and what evidence resolves it, distinguishes sibling folders that are each plausible in isolation, and recognizes when to abstain rather than force an assignment.
It must also improve during use from the only signal the user provides, an accept or reject on each suggestion, with no data to train on in advance.
No single classifier or LLM call handles all of this at once, so we cast routing as an agent that maintains state across steps and acts on each in turn.

\section{System Design}
\label{sec:method}

\subsection{Overview}
\label{sec:method:overview}
\begin{figure}[!htbp]
  \centering
  \includegraphics[width=0.95\linewidth]{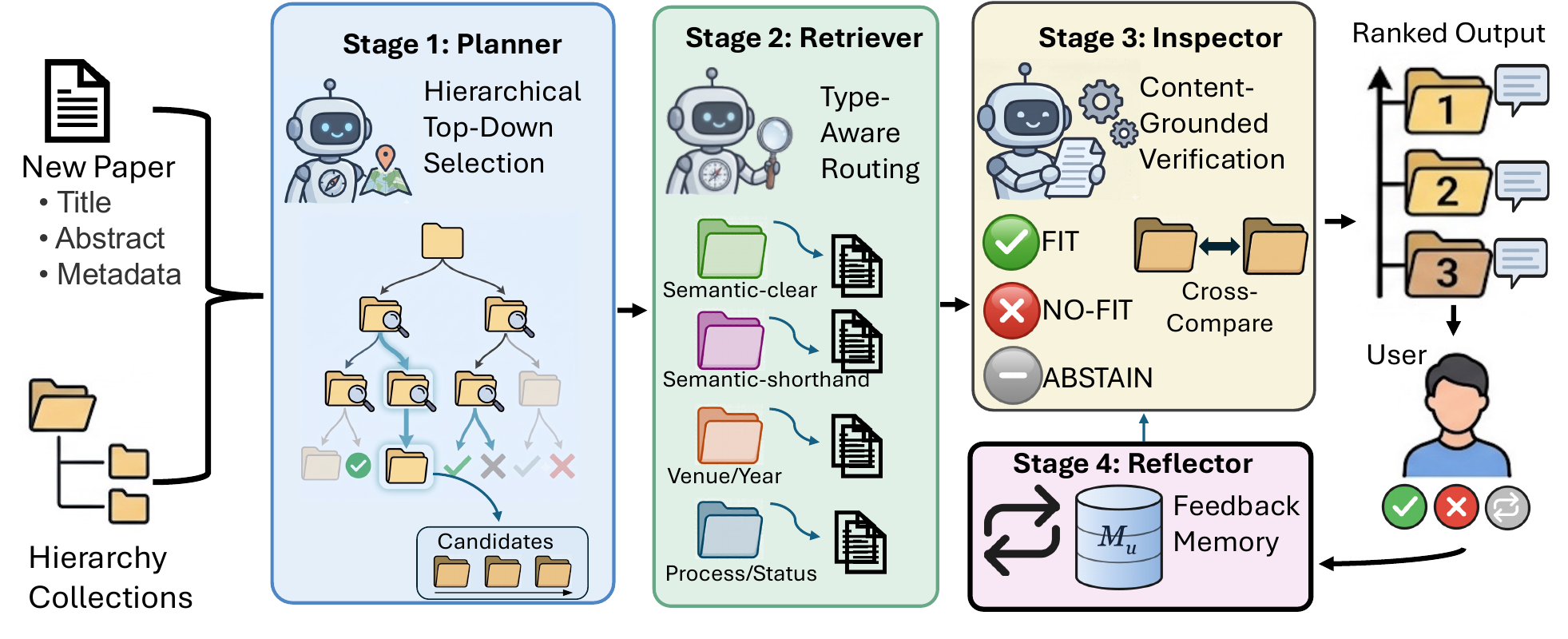}
  \caption{PaperRouter-Agent: a training-free four-stage agent that routes an incoming paper $p$ into the user's hierarchy $H_u$. The Planner short-lists candidate folders by top-down traversal. The Retriever types each folder and routes it, sampling member papers for topical folders while matching venue/year folders directly by metadata. The Inspector then reads the sampled evidence to decide \textsc{Fit}/\textsc{No-Fit}/\textsc{Abstain} and cross-compares survivors into a ranked list $R$. Accept/reject/override feedback in memory $M_u$ drives the Reflector, which injects relevant past rejections back into the Inspector.}
  \label{fig:architecture}
\end{figure}

PaperRouter-Agent decomposes PHPR into four LLM-driven stages (Planner, Retriever, Inspector, and Reflector) run in sequence, with the user's feedback looping back into later decisions (Fig.~\ref{fig:architecture}). 
It is designed around the task demands set out in Section~\ref{sec:problem}: reading each folder by its own contents, choosing evidence by folder type, weighing competing candidates, and learning from the user's streaming accept/reject, none of which a flat single-shot classifier provides. 
We also let the agent spend LLM reasoning only where reading content actually decides the answer. 
For example, a folder named \emph{CVPR} is settled by the paper's venue field and never needs its contents read. We describe each stage in turn, showing how each addresses these demands.

\subsection{Planner: Hierarchical Top-Down Selection}
\label{sec:method:planner}

The Planner narrows the user's hierarchy to a small set of candidate folders, so the later stages reason over a handful rather than the whole library. 
Serializing every leaf path into one prompt, as the single-shot baseline does, spreads attention across many similar paths and flattens the tree into a list of strings, discarding the parent--child structure that distinguishes same-named leaves under different parents.

The Planner instead traverses the hierarchy top-down. 
At each level, the model ranks only the five to ten folders directly under the current node and descends into the most promising branches. 
Top-down traversal can commit to a wrong branch early, a known weakness of hierarchical classifiers.
Our Planner therefore keeps every branch whose score is close to the best, not just the top one. 
Ambiguous siblings are passed to the Inspector, which separates them by content (Section~\ref{sec:method:inspector}). 
The Planner chooses what to consider, not the final routing.

\subsection{Retriever: Type-Aware Evidence and Routing}
\label{sec:method:retriever}

The Retriever decides, for each candidate folder, what evidence its decision needs and whether it needs the Inspector at all. Although Section~\ref{sec:problem} lists several folder types, for routing they reduce to two: topical folders, decided by reading what their papers are about, and organizational ones such as venue or year, fixed by the incoming paper's metadata.

The Retriever turns this distinction into a routing rule. 
It types each folder once, from the name and a quick look at its members, and caches the result.
Venue and year folders are matched directly against the paper's metadata, skipping the Inspector. 
This both saves the cost of reading content and surfaces folders a purely content-based method would never propose, the failure that sinks the single-shot baseline (Section~\ref{sec:eval:user-study}). 
Topical folders instead have representative members sampled for the Inspector. Process and status folders such as \emph{To Read} fit neither route and remain the residual hard case.

\subsection{Inspector: Content-Grounded Verification with Abstention}
\label{sec:method:inspector}

The Inspector decides whether the incoming paper belongs in a candidate folder by reading the folder's own member papers, not its name. 
It is the core of PaperRouter-Agent and the mechanism we evaluate directly (Section~\ref{sec:eval:lamp2}). 
For each topical candidate from the Planner, it takes the paper $p$, the folder name, and the sampled members, and returns \textsc{Fit} with a short rationale, \textsc{No-Fit}, or \textsc{Abstain}, in two steps: a per-folder check, then a cross-comparison of the survivors (pseudocode in Appendix~\ref{supp:algorithm}).

In the per-folder check, the Inspector judges each candidate on its own members, and abstains when they are too few or too mixed to decide rather than forcing a weak \textsc{Fit}. Shown a folder named \emph{anomaly} whose members are all 2D image defects, for instance, it returns \textsc{No-Fit} on a 3D point-cloud paper instead of matching the shared word.

The cross-comparison separates sibling folders that each look like a fit on their own. 
The Inspector compares all \textsc{Fit} candidates together and asks the model to name the signal between them. For \emph{2D/Detection} (image defects) and \emph{3D/Detection} (surface scans), a paper on 3D surface scans goes to the latter by modality, rather than being averaged into two near-identical scores. These are the ambiguous siblings the Planner kept for this step.

\subsection{Reflector: Feedback Memory}
\label{sec:method:reflector}

The Reflector lets the agent improve from use without retraining. When the user rejects a suggested folder or files a paper elsewhere, it records that decision in a per-user memory store $M_u$. On a later paper, it retrieves from $M_u$ the rejections relevant to a candidate folder (those whose papers are similar to the current one) and adds them to that folder's Inspector prompt. A folder the user has repeatedly declined for this kind of paper is then judged more cautiously, not penalized outright.

\begin{table}[!htbp]
\vspace{-10pt}
  \centering
  \caption{One representative baseline failure per folder type. The single-shot baseline routes by folder \emph{name} and never reads its contents.}
  \label{tab:userstudy-badcases}
  \small
  \setlength{\tabcolsep}{0pt}
  \begin{tabularx}{\linewidth}{@{}>{\bfseries\raggedright\arraybackslash}p{0.21\linewidth} @{\hspace{1.2em}} >{\raggedright\arraybackslash}X@{}}
    \toprule
    \normalfont\emph{Folder type} & \normalfont\emph{Representative baseline failure} \\
    \midrule
    semantic-clear & \emph{Multi-level MoE for Multimodal Entity Linking} $\rightarrow$ \fld{RAG/Multimodal}; should be \fld{MoE}. Matched the \fld{Multimodal} leaf under the wrong parent. \\ \addlinespace
    semantic-shorthand & \emph{Certainty in Uncertainty: Reasoning over Uncertain KGs} $\rightarrow$ unrouted; should be \fld{UKG}. Read the shorthand literally, missing U\,$=$\,uncertain. \\ \addlinespace
    process/status & \emph{SAM 3: Segment Anything with Concepts} $\rightarrow$ filed by topic; should be \fld{To Read}. Membership is reading intent, not content. \\ \addlinespace
    venue/year & \emph{LISA: Reasoning Segmentation via LLM} $\rightarrow$ filed by topic; should be \fld{CVPR}/\fld{2024}. No link from content to a venue or year. \\
    \bottomrule
  \end{tabularx}
\vspace{-12pt}
\end{table}

\section{Experiments}
\label{sec:eval}

\subsection{Implementation details}

Both systems share the same backbone LLM, \texttt{gpt-4o-mini}, so every comparison reflects the routing strategy rather than the underlying model. 
The core logic is written in Python. For the user study and real-world use, we implemented a deployment-ready Zotero integration (JavaScript and Python) that participants could run inside their own libraries; Appendix~\ref{supp:system} shows this deployed integration in use.

We build two frameworks on this backbone. 
The first is the \emph{single-shot baseline}, shown as \emph{PaperRouter (single-shot)}, or simply \emph{baseline}, in the results tables. 
It feeds the user's entire current hierarchy, together with the incoming paper's title and abstract, to the LLM in a single prompt and asks it to return the target folder(s) in one call.
Its failure cases on real libraries, together with the formative feedback from participants (Section~\ref{sec:eval:user-study}), motivated the second framework, \textbf{PaperRouter-Agent}, the four-stage agent of Section~\ref{sec:method} that grounds each decision in folder members.

We keep the agent's settings deliberately light. 
The Planner ranks the folders directly under each node (5-10, 5 by default) and keeps every branch scoring near the top.
The Retriever types each folder once from its name and a sample of its members, caches the result, routes venue and year folders by metadata, and samples representative members of topical folders for the Inspector. 
The Inspector returns \textsc{Fit}/\textsc{No-Fit}/\textsc{Abstain}, abstaining when a folder's members are too few or too mixed to decide, and by default returns at most 3 candidate folders for the user to choose from. 
The Reflector stores accept/reject feedback in a per-user memory and injects the most similar past rejections into the relevant Inspector prompt. 

\subsection{User Study on Real Libraries}
\label{sec:eval:user-study}

\textbf{Setup.} We ran a two-round within-subjects study on five research students' own Zotero libraries: each filed 20 papers they normally read into their own collections, once under the deployed single-shot baseline and once under the full agent (full protocol in Appendix~\ref{supp:protocol}). So that Recall is meaningful, each participant recorded the correct folder(s) for every paper (the gold set $R^\star$) \emph{independently} of what the system returns: they accepted the correct returned candidates and named any correct folder the system omits. Without this, a folder the system never surfaces (typically a venue or year folder) could never be counted as a miss.

\textbf{Evaluation metric.} Given the gold set $R^\star$ and the system's ranked candidate list, we report \textbf{Recall@$K$}$=|\text{top-}K \cap R^\star|/|R^\star|$ for $K\!\in\!\{1,3\}$, averaged over papers. The full protocol and the prompt templates are in Appendices~\ref{supp:protocol} and~\ref{supp:impl}.

Table~\ref{tab:userstudy-badcases} gives one representative failure per folder type, all from a single cause: the baseline matches the folder \emph{name} and never reads its contents.
Reading folder contents recovers most of this gap (Table~\ref{tab:userstudy}). The baseline handles the semantic-clear majority but collapses on the non-semantic types, falling to Recall@1 $0.09$ on venue/year. The full agent lifts overall Recall@1 from $0.39$ to $0.61$ and Recall@3 from $0.57$ to $0.83$, with the largest gains where the baseline fails (venue/year $0.09\to0.50$) and every participant improving. Process and status folders such as \emph{To Read} remain the residual hard case.

\begin{table}[!htbp]
\vspace{-5pt}
  \centering
  \caption{User study on five participants' real Zotero libraries: Recall@1 and Recall@3 by folder type, single-shot baseline vs.\ full agent; $n$ is the number of papers per folder type (100 total).}
  \label{tab:userstudy}
  \begin{tabular*}{\linewidth}{@{\extracolsep{\fill}} l c c c c c @{}}
    \toprule
     & & \multicolumn{2}{c}{Recall@1} & \multicolumn{2}{c}{Recall@3} \\
    \cmidrule(lr){3-4}\cmidrule(lr){5-6}
    Folder type & $n$ & Base & Ours & Base & Ours \\
    \midrule
    Semantic-clear     & 52  & 0.62 & 0.72 & 0.78 & 0.91 \\
    Semantic-shorthand & 18  & 0.20 & 0.52 & 0.47 & 0.79 \\
    Process/status     & 12  & 0.12 & 0.44 & 0.31 & 0.70 \\
    Venue/year         & 18  & 0.09 & 0.50 & 0.23 & 0.74 \\
    \midrule
    \rowcolor{oursrow}
    \textbf{Overall}   & 100 & \textbf{0.39} & \textbf{0.61} & \textbf{0.57} & \textbf{0.83} \\
    \bottomrule
  \end{tabular*}
  \vspace{-5pt}
\end{table}

The same pattern holds on LaMP-2 (Section~\ref{sec:eval:lamp2}), where content-grounded inspection again helps most on the subtle classes. A longitudinal study with online feedback, which would exercise the Reflector, is left to future work.

\subsection{LaMP-2 (Personalized Movie Tagging)}
\label{sec:eval:lamp2}

We evaluate on \textbf{LaMP-2}~\cite{salemi2024lamp}, the public benchmark whose per-user structure is closest to PHPR.
The task is to predict which of fifteen fixed tags a user assigns to a movie, given its plot description and that user's history of past \emph{(description, tag)} taggings. 
For each tag, the user's past taggings give the movies they already assigned it, and the Inspector reads these just as it reads the papers inside a folder.

Our evaluation here targets the Inspector, the content-grounding mechanism at the core of the framework, rather than the full pipeline. 
A public benchmark has none of the ambiguous, non-topical folder names, such as private shorthand or venues, that the Planner, Retriever, and Reflector exist to resolve, so those stages have nothing to act on and are evaluated separately in the user study (Section~\ref{sec:eval:user-study}).

\begin{table}[!htbp]
  \centering
  \caption{LaMP-2~\cite{salemi2024lamp}: comparison among \emph{zero-shot} (no in-context demonstrations) methods. Baseline numbers are as reported by Salemi et al.~\cite{salemi2024lamp}; inline citations mark each baseline's underlying model and retriever. Our PaperRouter runs under \texttt{gpt-4o-mini} in two modes: a single-shot classifier, and PaperRouter-Agent (mean over three seeds).}
  \label{tab:lamp2-results}
  \begin{tabular*}{\linewidth}{@{\extracolsep{\fill}} l c c @{}}
    \toprule
    Method & Accuracy & Macro-F1 \\
    \midrule
    \multicolumn{3}{@{}l}{\emph{No retriever}}\\
    \quad FlanT5-XXL~\cite{chung2022flant5}                       & 0.365 & 0.308 \\
    \quad GPT-3.5~\cite{ouyang2022instructgpt}                    & 0.408 & 0.314 \\[2pt]
    \multicolumn{3}{@{}l}{\emph{With retriever}}\\
    \quad GPT-3.5~\cite{ouyang2022instructgpt} + Contriever~\cite{izacard2022contriever} & 0.508 & 0.457 \\
    \midrule
    \multicolumn{3}{@{}l}{\emph{Ours (training-free, no retriever)}}\\
    \quad PaperRouter (single-shot)           & 0.445 & 0.364 \\
    \rowcolor{oursrow}
    \quad \textbf{PaperRouter-Agent}          & \textbf{0.515} & \textbf{0.454} \\
    \bottomrule
  \end{tabular*}
\end{table}

\begin{figure}[!htbp]
  \centering
  \includegraphics[width=\linewidth]{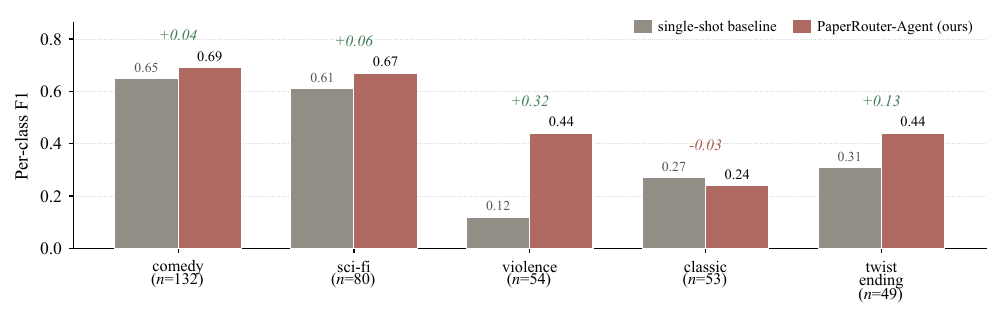}
  \caption{Per-class F1 on LaMP-2 (five most frequent tags, $n$ = count in dev), single-shot baseline vs.\ PaperRouter-Agent. Content-grounding helps most on subtle, content-determined tags such as \emph{violence} ($+0.32$) and \emph{twist ending}.}
  \label{fig:perclass}
\end{figure}

Content-grounding improves accuracy from 44.5\% to 51.5\% ($+7.0$ points, $+9.0$ macro-F1), bringing a training-free model with no separate retriever up to the level of a retrieval-augmented LLM (GPT-3.5\,+\,Contriever; Table~\ref{tab:lamp2-results}). The gains concentrate on subtle tags whose name alone is uninformative, such as \emph{violence} and \emph{twist ending} (Fig.~\ref{fig:perclass}).

\subsection{Cost and Latency Analysis}
\label{sec:eval:cost}

\begin{figure}[!htbp]
  \centering
  \includegraphics[width=0.9\linewidth]{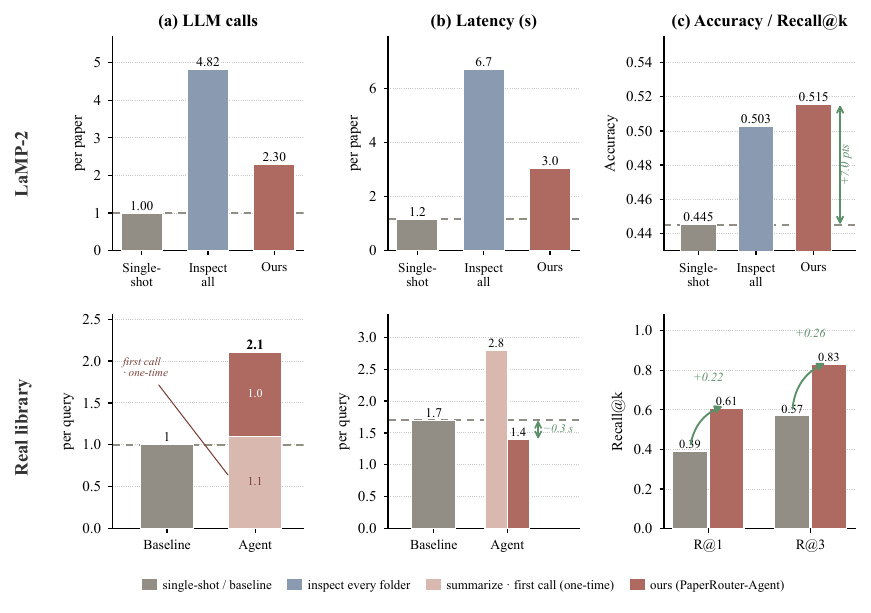}
  \caption{\textbf{Cost, latency, and quality} (\texttt{gpt-4o-mini}). Columns are (\textbf{a})~LLM calls, (\textbf{b})~latency, and (\textbf{c})~quality. The two rows are LaMP-2 (per paper) and real libraries (per query). The lighter bar is the one-time per-folder summary, and the gray dashed line marks the baseline.}
  \label{fig:cost-latency}
\end{figure}

PaperRouter-Agent normally calls the LLM more than once per paper, so it can add latency over a single call. 
We report latency and call counts on \texttt{gpt-4o-mini}, averaged over LaMP-2 and the five participants' real sessions (Fig.~\ref{fig:cost-latency}), against two references: the \emph{single-shot} baseline (one call, but a long prompt consisting of most of the hierarchy) and \emph{inspect-all} (reading every candidate folder's members on every decision). 
Our agent instead reads a folder's members only when its name needs interpreting, and caches the result.

On LaMP-2, where every paper is routed from scratch, the agent uses 2.3 calls and 3.0 seconds per paper, about $2.5\times$ single-shot but half of inspect-all, and it is the most accurate of the three. 
A real library front-loads the cost instead: the first routing into a folder runs a one-time summary of its members and is slow, but later queries run even faster than single-shot, since a folder settled by metadata such as a year is matched with no further reading while single-shot still pays for its long prompt. 
Steady-state routing then costs about one call and 1.4 seconds, below the baseline's 1.7 seconds, while raising Recall@1 by 0.22 and Recall@3 by 0.26.

\section{Discussion}
\label{sec:discussion}

As paper routing is inherently personalized, we evaluate it the way the setting calls for, on researchers' own libraries with their real accept and reject feedback rather than on a fixed offline corpus. 
Further details of the agent, the full user-study protocol, qualitative examples of the deployment, and the prompt templates are provided in the appendix. 
What this design cannot yet settle is a mechanism like the Reflector, which is built to improve as feedback accumulates over repeated use: a single-session study points to that effect but does not measure it. The task also has no large public benchmark, so our offline comparison falls back on LaMP-2 as the closest available proxy.

Building such a benchmark is itself a direction worth pursuing: a larger, openly reviewable dataset and protocol for this setting would let personalized routing, and in particular the long-term gains the Reflector is designed for, be measured properly. 
We also plan to keep refining the system on the engineering side, where much of the routing cost and latency can still be reduced.

\section{Conclusion}
\label{sec:conclusion}

We introduced \emph{personalized hierarchical paper routing} (PHPR), the problem of routing a paper into a researcher's own evolving hierarchy. Classical hierarchical text classification does not cover it, because a personal library is a folksonomy rather than the fixed, shared taxonomy those methods assume. Our system, PaperRouter-Agent, is training-free and routes by reading the papers already inside a folder instead of matching its name. On five researchers' real Zotero libraries it raises Recall@1 from 0.39 to 0.61 and Recall@3 from 0.57 to 0.83, with the largest gains where name matching breaks down, such as venue and year folders. On the LaMP-2 proxy benchmark the same content-grounded inspection lifts accuracy from 44.5\% to 51.5\% with a 9.0-point macro-F1 gain, matching a retrieval-augmented LLM while using no separate retriever and no fine-tuning. Grounding a model's decisions in the user's own data, rather than in folder names or external definitions, is a principle we expect to carry over to other personal-organization tasks.

\bibliographystyle{splncs04}
\bibliography{paperrouter}

\clearpage
\appendix

\begin{center}
  {\LARGE\bfseries Appendix}\\[4pt]
  \rule{0.45\linewidth}{0.4pt}
\end{center}
\vspace{1em}

\noindent This appendix collects material referenced from the main paper:
qualitative routing examples from the deployed integration
(Appendix~\ref{supp:system}), the full user-study protocol
(Appendix~\ref{supp:protocol}), the Inspector pseudocode
(Appendix~\ref{supp:algorithm}), and the prompt templates and hyperparameters
(Appendix~\ref{supp:impl}).

\section{Qualitative Routing Examples from the Deployed Integration}
\label{supp:system}
We show qualitative results from the deployed integration used in our user study
(Section~\ref{sec:eval:user-study}): the interface, several routing decisions on real
papers with the per-folder \textsc{Fit}/\textsc{No-Fit}/\textsc{Abstain} verdicts
and confidences from the Inspector (Section~\ref{sec:method:inspector}), and one
failure case. The integration runs as a Zotero plugin inside each participant's own
library. In the figures, a folder name shown in magenta marks a collection the paper
is \emph{already} filed in, the \texttt{Conf.}\ value is the agent's confidence for
that folder, and the \texttt{Thrd.}\ control sets the pre-selection threshold.

\subsection{Configuration}
\label{supp:system:config}
\begin{figure}[H]
  \centering
  \includegraphics[width=\linewidth]{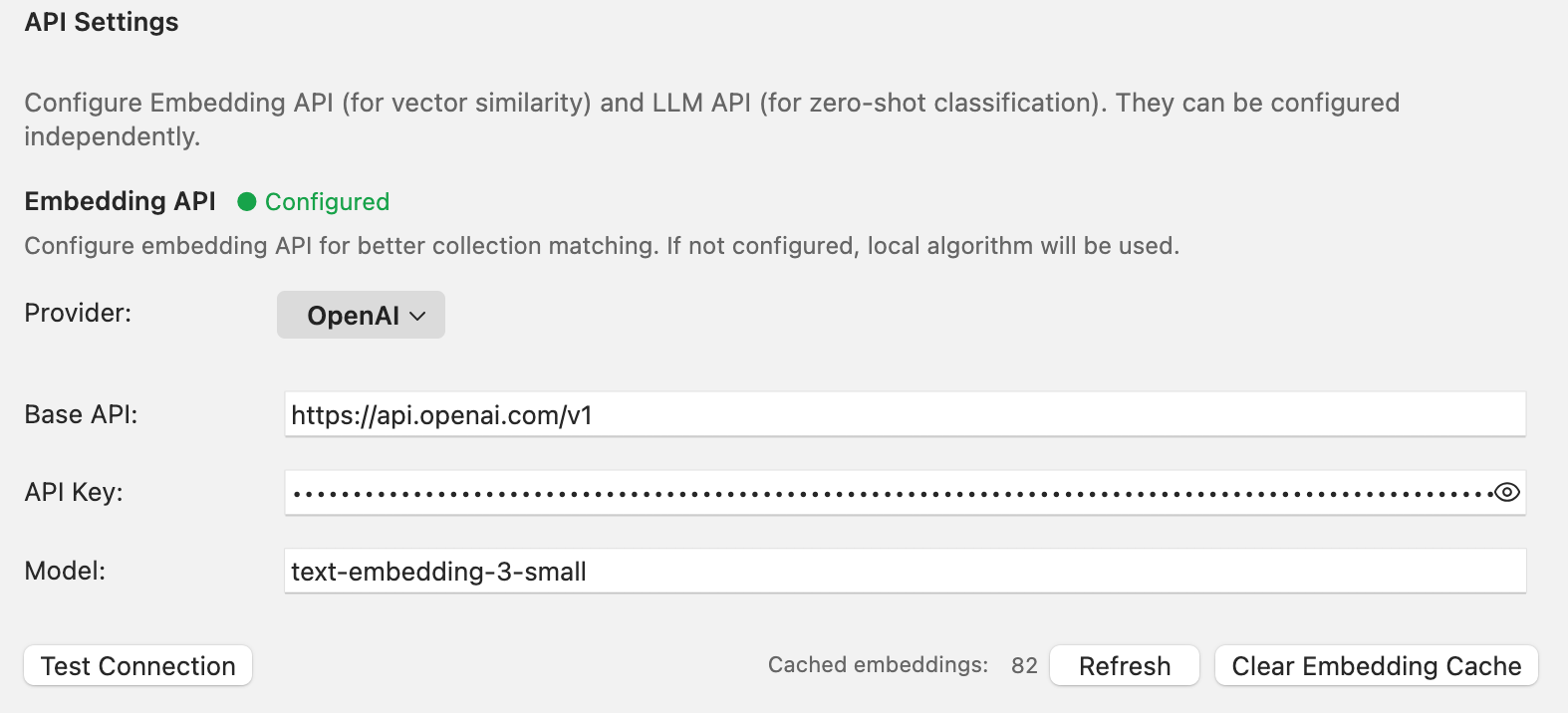}
  \caption{Configuration: the embedding API and the LLM API are set independently.}
  \label{fig:supp-setting}
\end{figure}

Figure~\ref{fig:supp-setting} shows the configuration panel. Two APIs are set
independently: an \emph{embedding} API for the vector-similarity steps (candidate
narrowing and the similarity-gated retrieval of past rejections in the Reflector),
and an \emph{LLM} API for zero-shot classification and the content-grounded
verification of the Inspector. Decoupling them lets the embedding and LLM models be
chosen independently, and cached embeddings are reused across sessions.

\subsection{Routing Output and Confidence}
\label{supp:system:confidence}
\begin{figure}[p]
  \centering
  \includegraphics[width=0.80\linewidth]{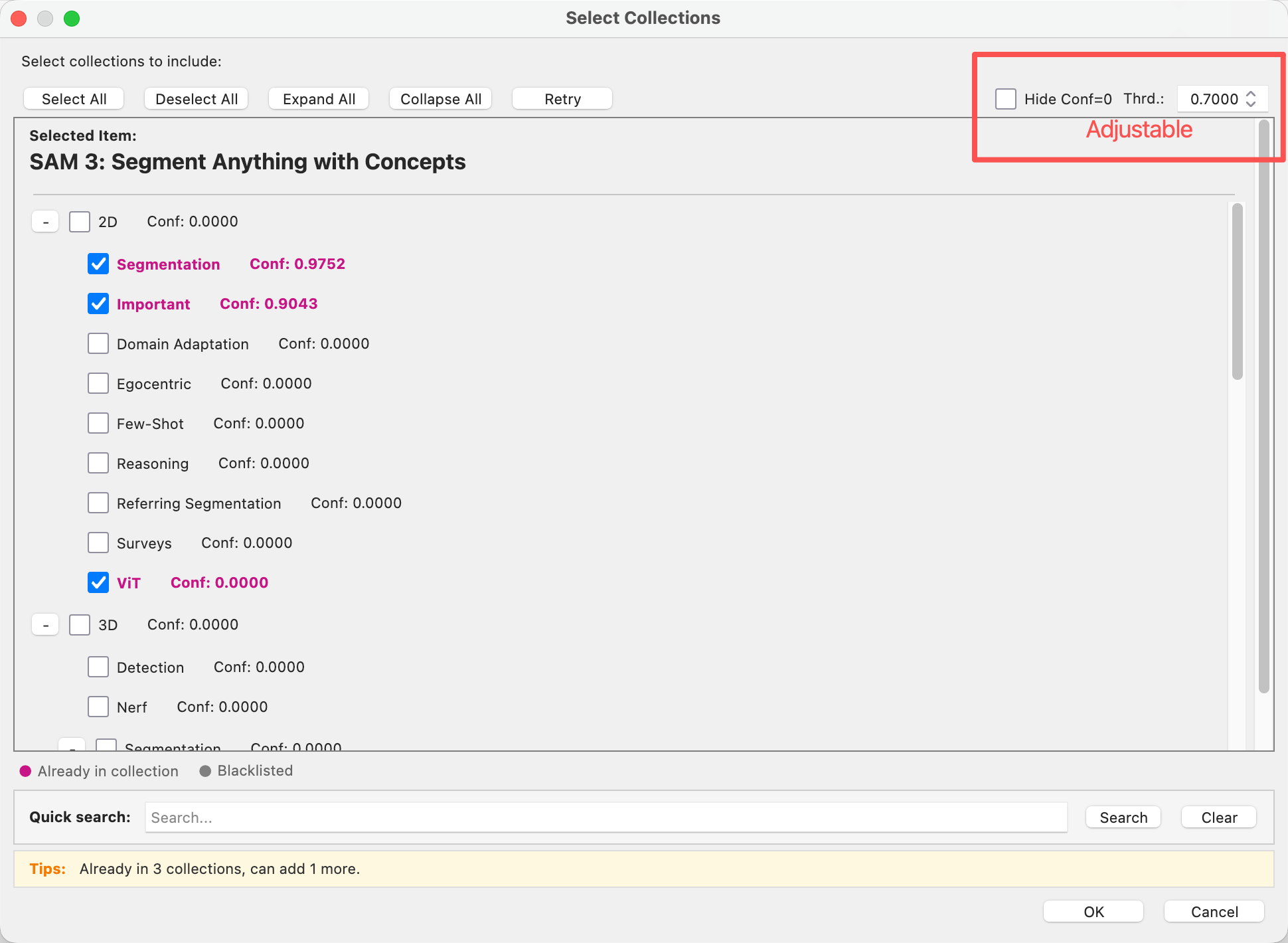}\\[8pt]
  \includegraphics[width=0.80\linewidth]{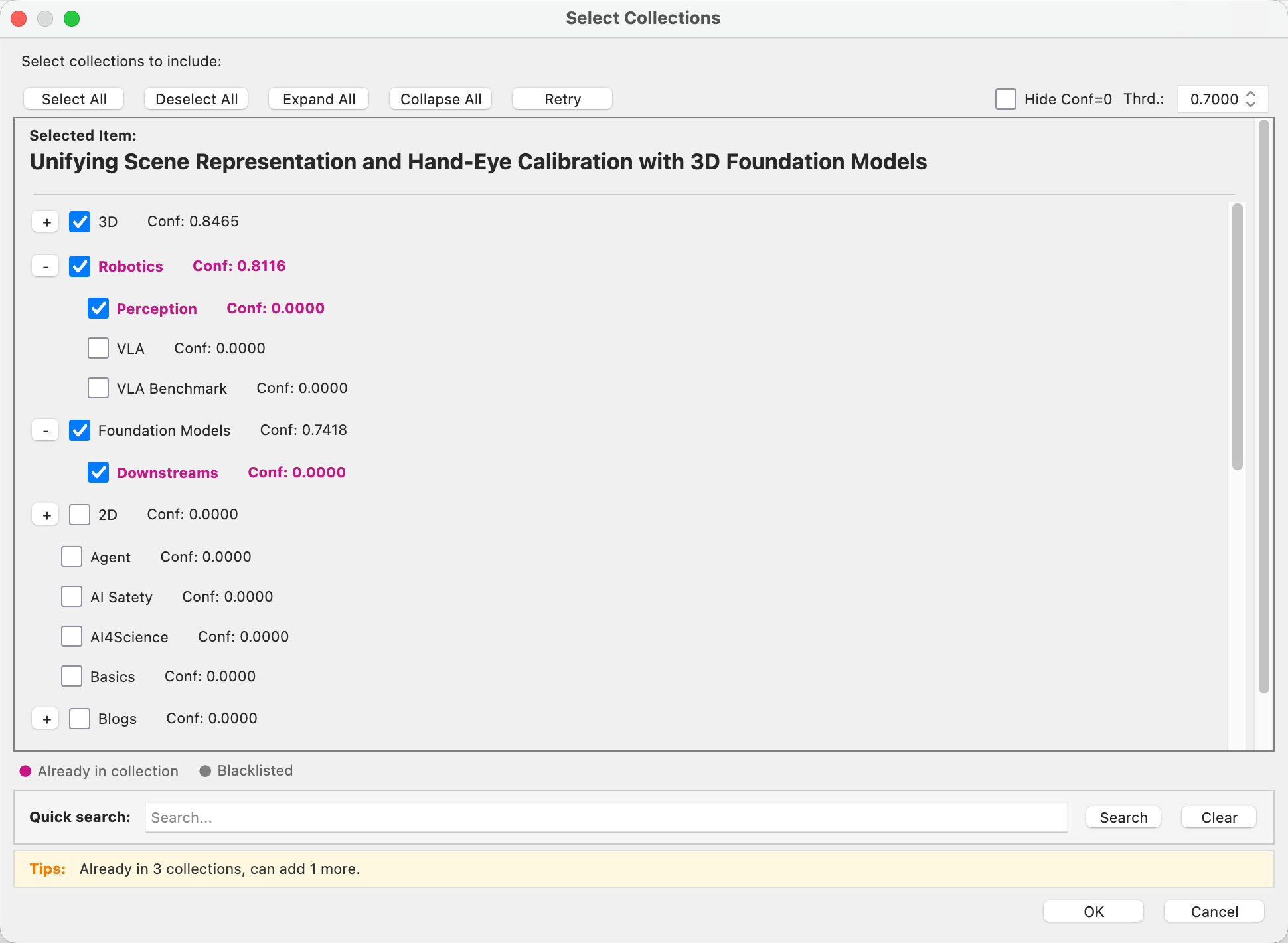}
  \caption{Routing output with per-folder confidence: (a)~\emph{SAM 3: Segment
  Anything with Concepts} and (b)~\emph{Unifying Scene Representation and Hand-Eye
  Calibration with 3D Foundation Models}.}
  \label{fig:supp-confidence}
\end{figure}

\begin{figure}[tbp]
  \centering
  \includegraphics[width=0.80\linewidth]{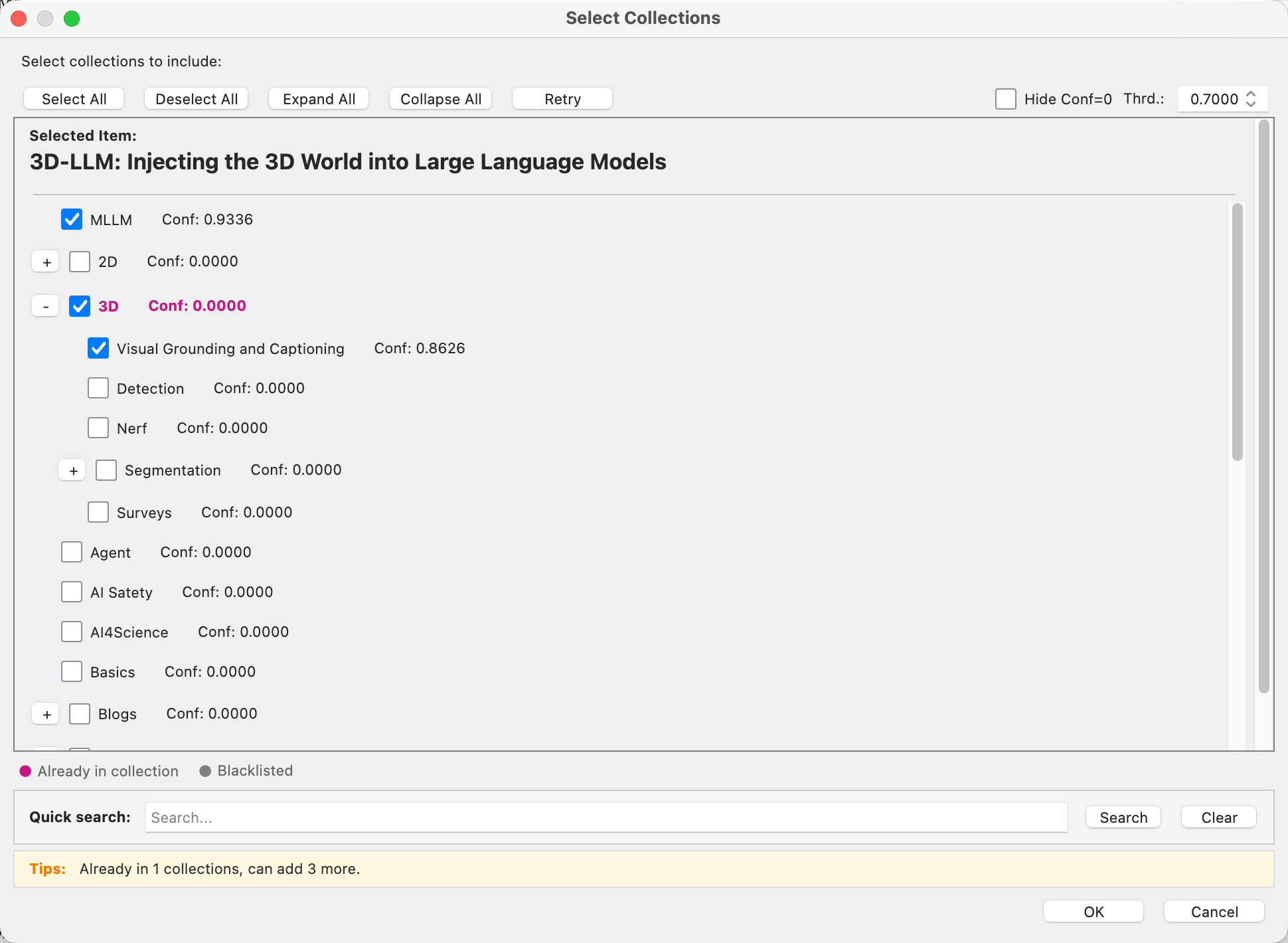}
  \caption{Routing output with per-folder confidence, continued: (c)~\emph{3D-LLM:
  Injecting the 3D World into Large Language Models}.}
  \label{fig:supp-confidence-cont}
\end{figure}

Figures~\ref{fig:supp-confidence} and~\ref{fig:supp-confidence-cont} show the routing
output on three papers of different types. The \emph{Select Collections} dialog lists
the candidate folders in the user's own hierarchy, each annotated with a confidence,
and pre-checks those above the adjustable threshold. The agent routes to folders at
different depths of the hierarchy, which a flat name match would miss.

\subsection{Content-Grounded Inspection}
\label{supp:system:inspection}
\begin{figure}[p]
  \centering
  \includegraphics[width=0.80\linewidth]{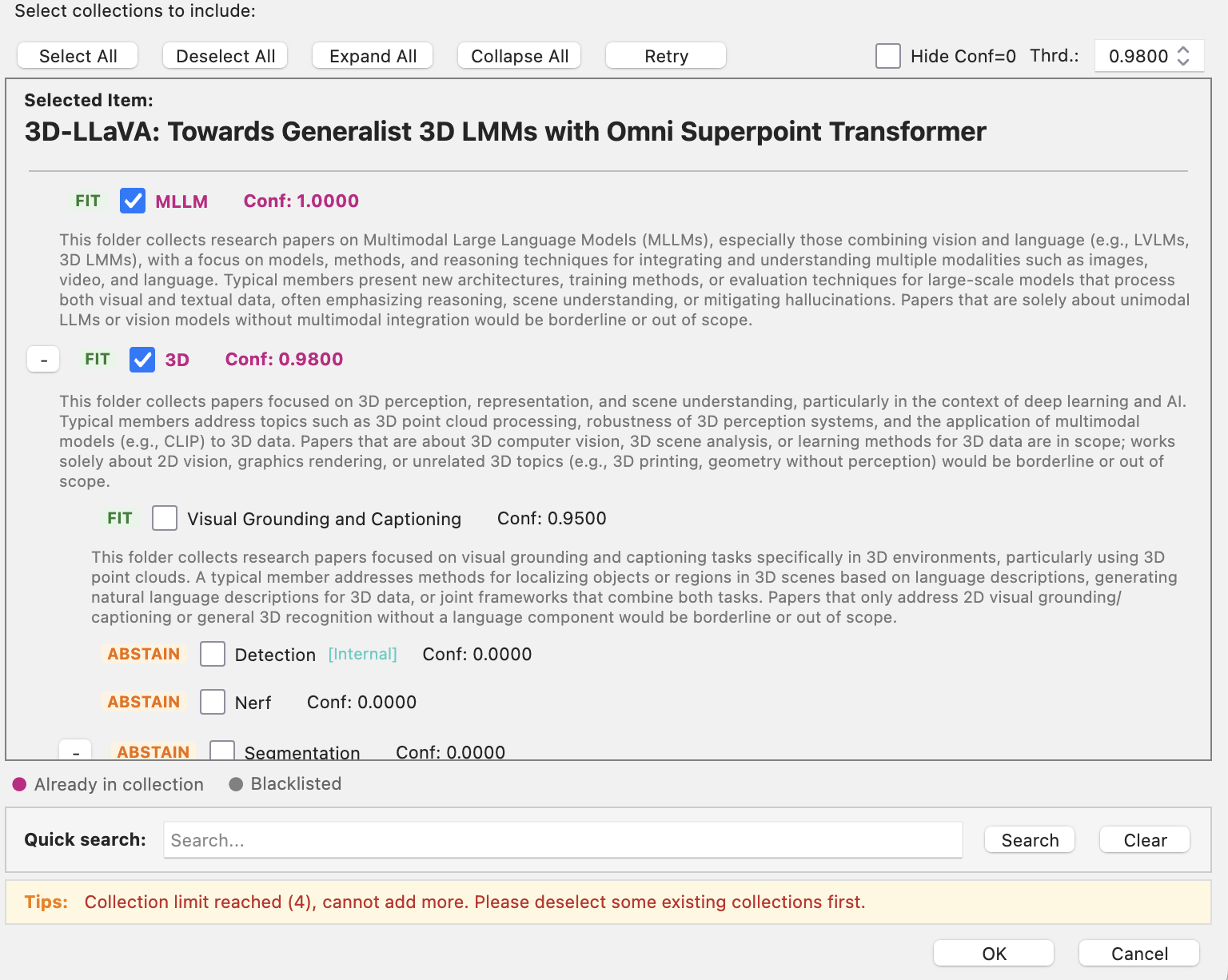}\\[6pt]
  \includegraphics[width=0.80\linewidth]{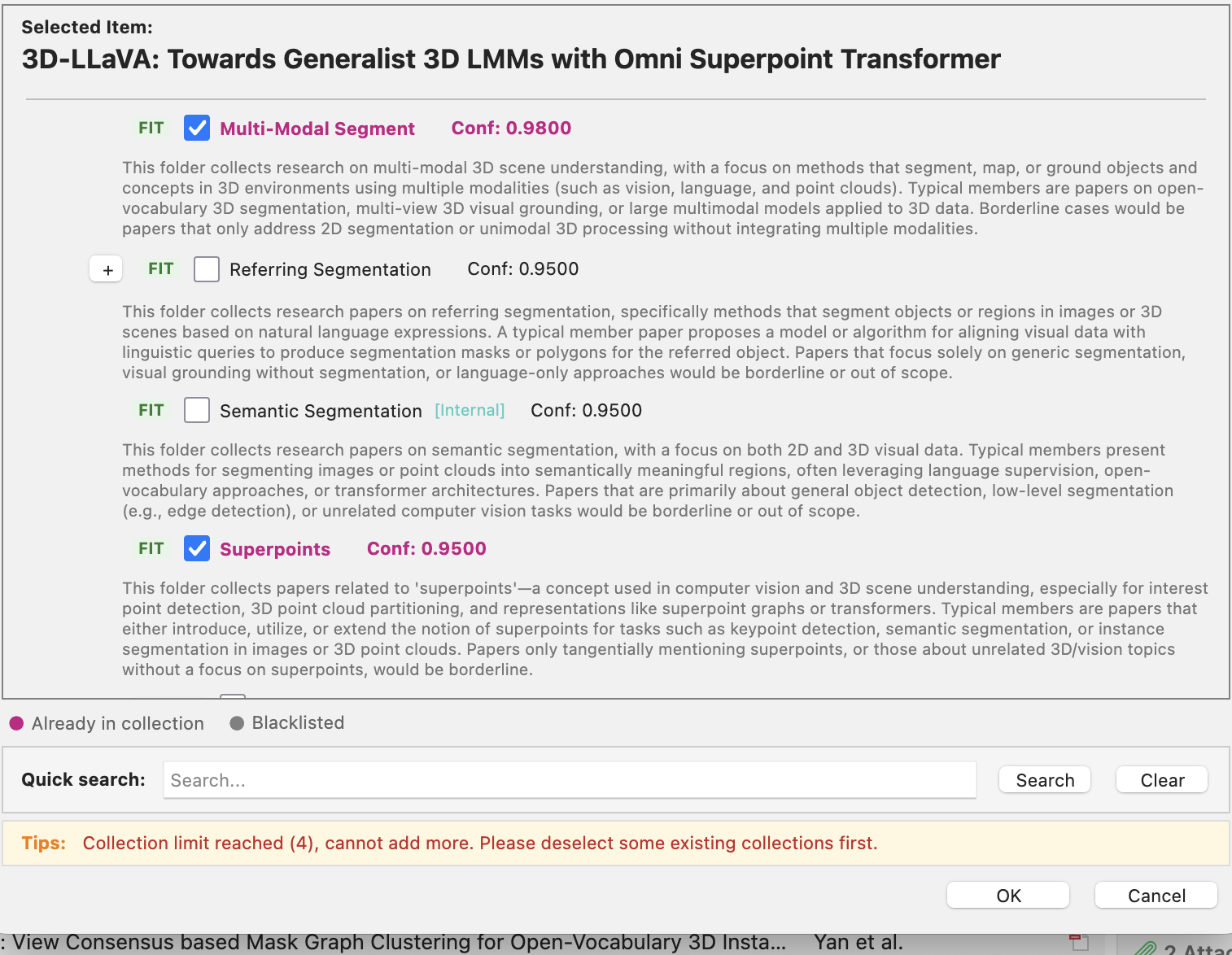}
  \caption{Content-grounded inspection for \emph{3D-LLaVA} (scroll positions 1--2 of
  three): each candidate folder shows a \textsc{Fit}\,/\,\textsc{No-Fit}\,/\,%
  \textsc{Abstain} verdict, a confidence, and the scope inferred from its members.}
  \label{fig:supp-3dllava}
\end{figure}

\begin{figure}[tbp]
  \centering
  \includegraphics[width=0.80\linewidth]{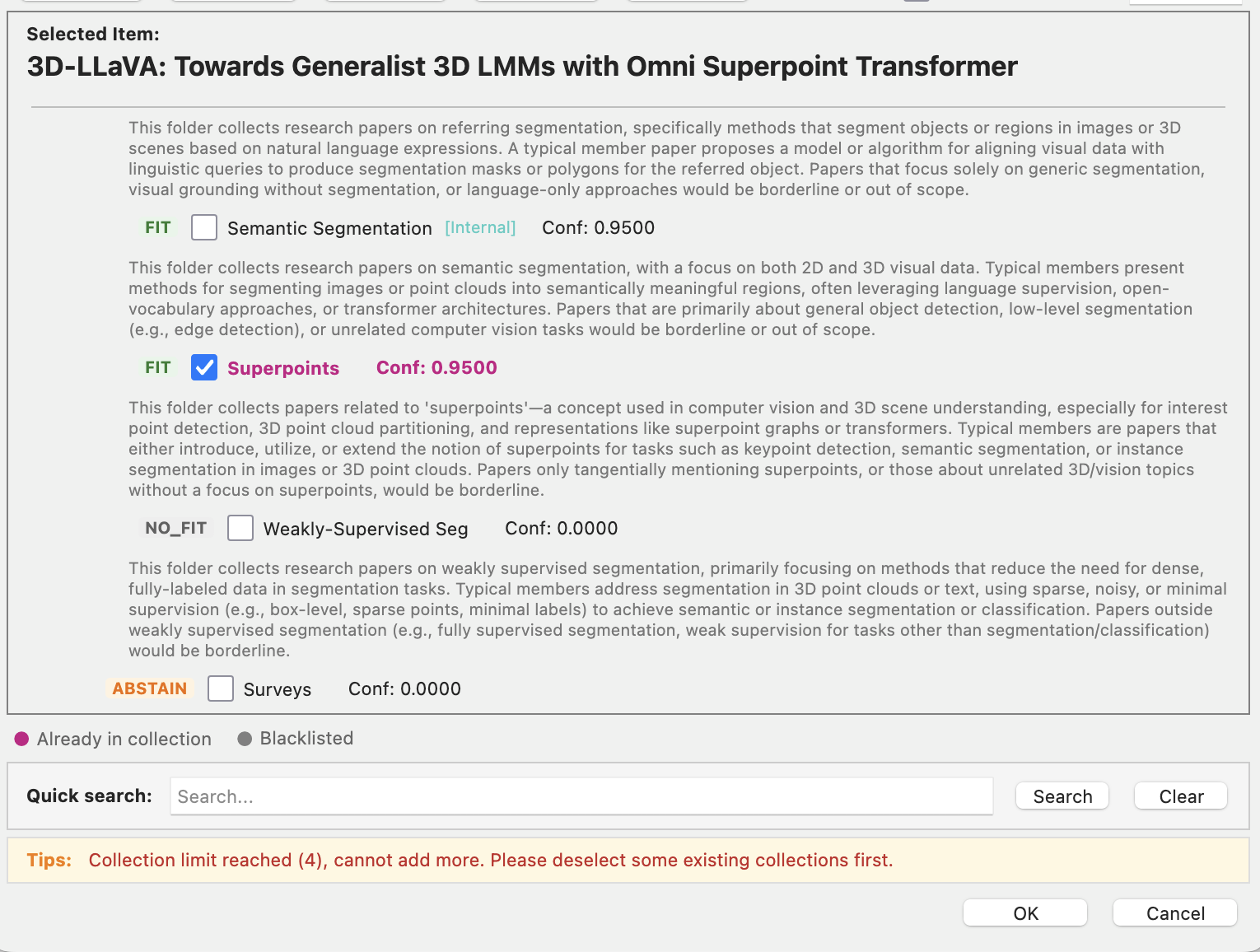}
  \caption{\emph{3D-LLaVA}, continued (scroll position 3 of three).}
  \label{fig:supp-3dllava-cont}
\end{figure}

Figures~\ref{fig:supp-3dllava} and~\ref{fig:supp-3dllava-cont} show the Inspector's
content-grounded verification for \emph{3D-LLaVA}. Each candidate folder carries a
verdict (\textsc{Fit}, \textsc{No-Fit}, or \textsc{Abstain}), a confidence, and the
scope the Inspector inferred from the folder's members. The Inspector abstains when
those members are too few or too mixed to decide, rather than forcing a weak
\textsc{Fit}.

\begin{figure}[p]
  \centering
  \includegraphics[width=0.84\linewidth]{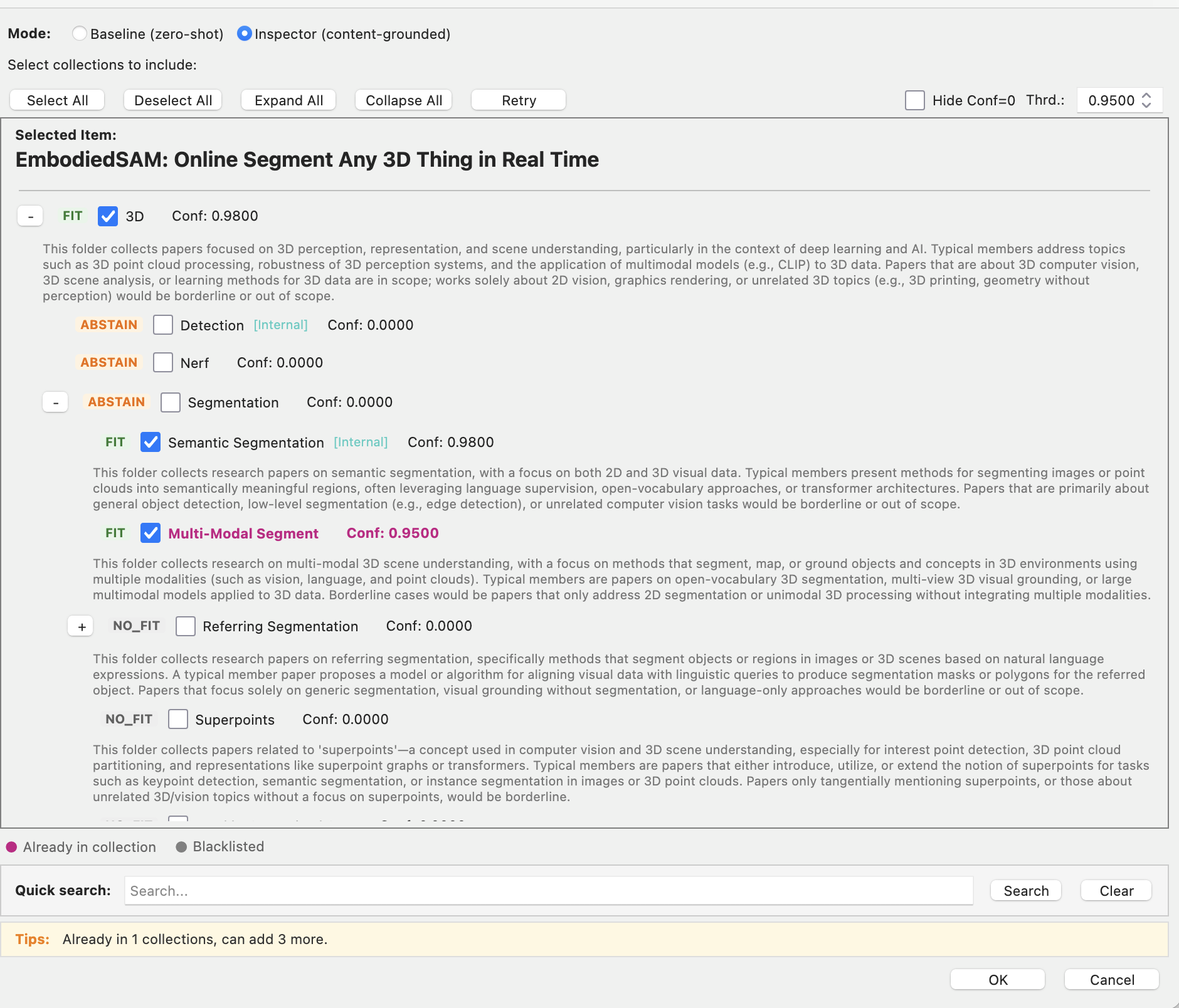}\\[8pt]
  \includegraphics[width=0.84\linewidth]{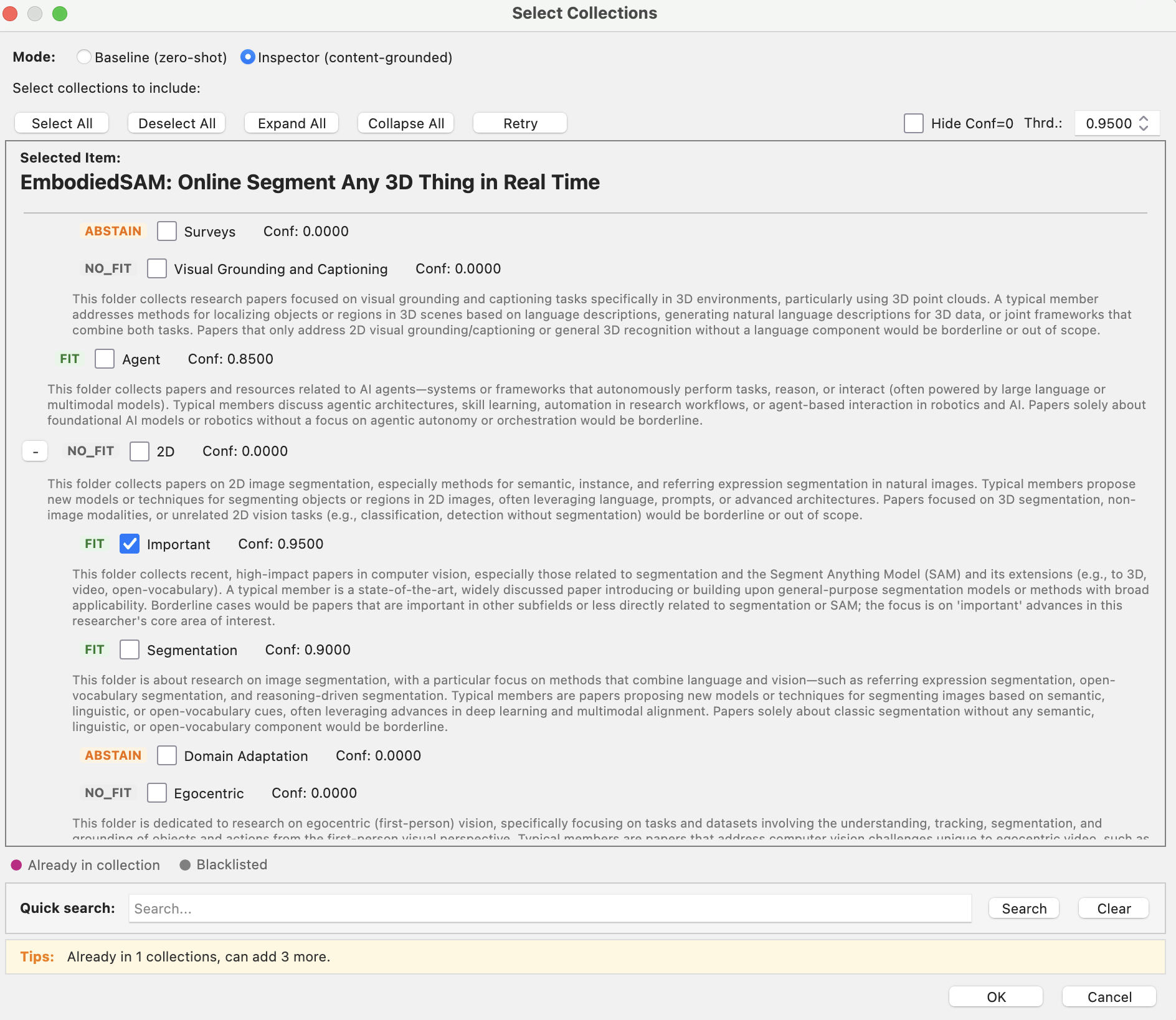}
  \caption{Content-grounded inspection for \emph{EmbodiedSAM} (two scroll positions).
  The \emph{Mode} toggle selects the single-shot \emph{Baseline} or the
  \emph{Inspector}.}
  \label{fig:supp-embodiedsam}
\end{figure}

Figure~\ref{fig:supp-embodiedsam} shows the same view for \emph{EmbodiedSAM},
including the \emph{Mode} toggle that switches between the single-shot baseline and
the Inspector. Each rejection there cites a scope mismatch rather than the absence
of a shared word.

\subsection{A Failure Case}
\label{supp:system:failure}
\begin{figure}[tb]
  \centering
  \includegraphics[width=\linewidth]{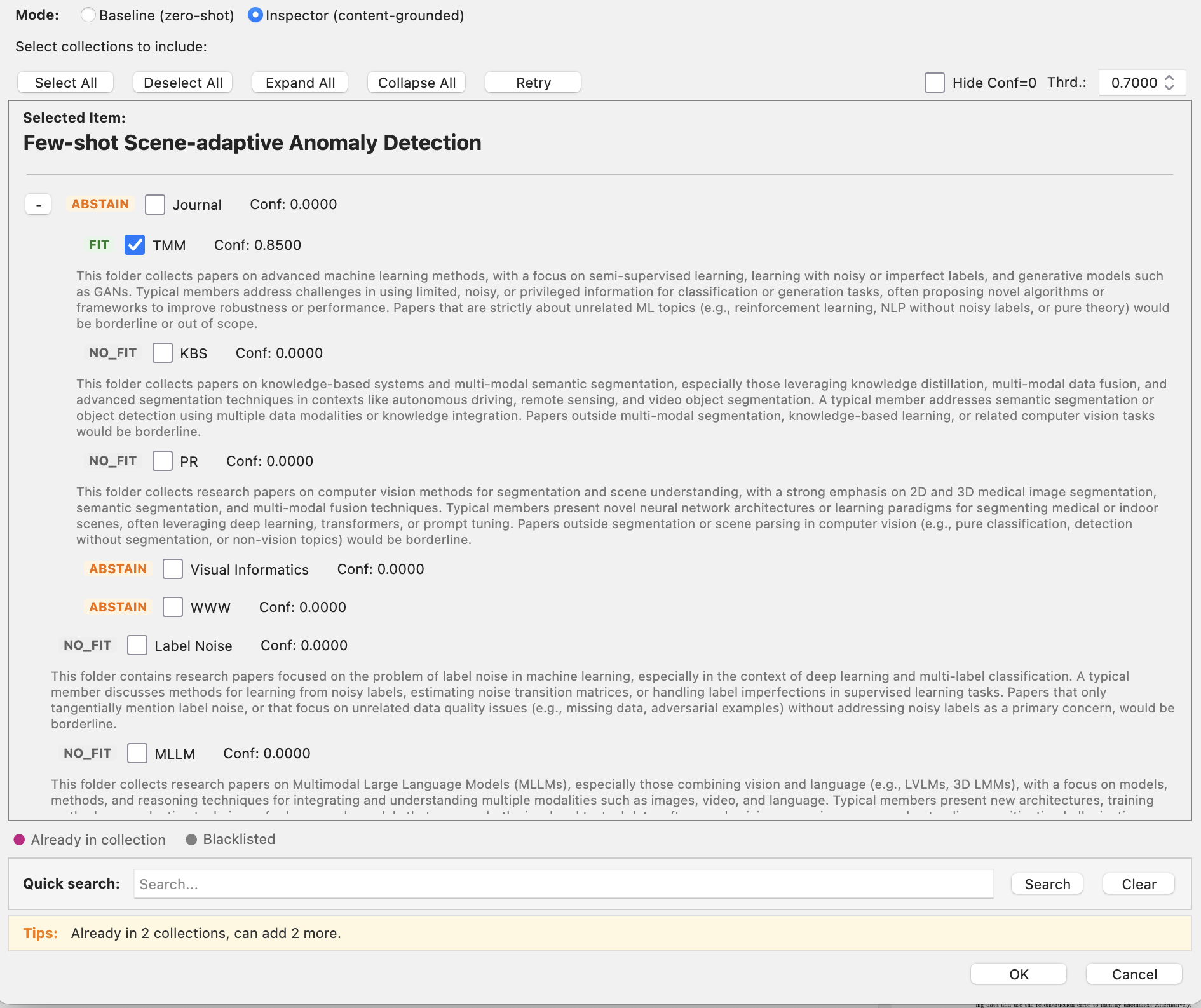}
  \caption{A failure case: \emph{Few-shot Scene-adaptive Anomaly Detection} is
  incorrectly accepted into \emph{TMM} (\textsc{Fit}, $0.85$).}
  \label{fig:supp-errorcase}
\end{figure}

Figure~\ref{fig:supp-errorcase} shows a failure of content grounding:
\emph{Few-shot Scene-adaptive Anomaly Detection} is incorrectly accepted into
\emph{TMM} (\textsc{Fit}, $0.85$) on a loose overlap with that folder's
heterogeneous members.

\section{User-Study Protocol}
\label{supp:protocol}
Table~\ref{tab:userstudy-protocol} gives the full protocol of the two-round study
summarized in Section~\ref{sec:eval:user-study}.

\begin{table}[!htbp]
  \centering
  \caption{Full protocol of the two-round user study.}
  \label{tab:userstudy-protocol}
  \small
  \begin{tabularx}{\linewidth}{@{}>{\bfseries}l X@{}}
    \toprule
    Participants & 5 research students (master's and PhD), each with an active multi-level Zotero library \\ \addlinespace
    Material     & 20 papers per participant, randomly drawn from those they normally read \\ \addlinespace
    Instruction  & \emph{``Randomly pick 20 papers you usually read and file each into your own hierarchy (collections).''} \\ \addlinespace
    Procedure    & For each paper the system returns a ranked candidate list; the participant accepts the correct candidates and names any correct folder the system omits, giving the gold folder set $R^\star$ \\ \addlinespace
    Conditions   & Round 1: single-shot baseline. \enspace Round 2: full PaperRouter-Agent (identical papers and protocol) \\
    \bottomrule
  \end{tabularx}
\end{table}

\section{Inspector Pseudocode}
\label{supp:algorithm}
Algorithm~\ref{alg:inspector} gives the pseudocode for the Inspector's two steps:
per-folder verification, then cross-comparison of the surviving \textsc{Fit}
candidates (Section~\ref{sec:method:inspector}).

\begin{algorithm}[ht]
\caption{Inspector: per-folder verification, then cross-comparison.}
\label{alg:inspector}
\begin{algorithmic}[1]
\Require incoming paper $p$; topical candidates $C$; for each $f \in C$, sampled members $M_f$ and relevant past rejections $R_f$ (from the Reflector)
\Ensure ranked list of fitting folders
\State $S \gets \emptyset$
\For{$f \in C$} \Comment{Step 1: per-folder verification}
  \State $(\ell, \rho) \gets$ \textsc{Verify}$(p, \mathrm{name}(f), M_f, R_f)$ \Comment{label $\ell$, rationale $\rho$}
  \If{$\ell = \textsc{Fit}$}
    \State $S \gets S \cup \{(f, \rho)\}$ \Comment{\textsc{No-Fit} and \textsc{Abstain} folders are dropped}
  \EndIf
\EndFor
\If{$|S| \le 1$}
  \State \Return $S$ \Comment{nothing to disambiguate}
\EndIf
\State \Return \textsc{Compare}$(p, S)$ \Comment{Step 2: joint ranking by the separating signal}
\end{algorithmic}
\end{algorithm}

\section{Prompts and Hyperparameters}
\label{supp:impl}
Table~\ref{tab:hparams} lists the hyperparameters and Appendix~\ref{supp:prompts}
gives the verbatim prompt templates. Placeholders are written \texttt{\{like\_this\}}.

\subsection{Hyperparameters}
\label{supp:hparams}

\begin{table}[H]
  \centering
  \caption{Routing hyperparameters.}
  \label{tab:hparams}
  \small
  \begin{tabular}{@{}lll@{}}
    \toprule
    Stage & Parameter & Value \\
    \midrule
    Backbone   & model                                   & \texttt{gpt-4o-mini} (via OpenRouter) \\
               & temperature                             & 0.0 \\
    \addlinespace
    Retriever  & members sampled per folder ($N$)        & 5 \\
               & sampling order                          & first-$N$ (deterministic) \\
               & member-abstract truncation              & 300 chars \\
    \addlinespace
    Inspector  & min.\ members to inspect a folder       & 3 \\
               & verdict labels                          & \textsc{Fit} / \textsc{No-Fit} / \textsc{Abstain} \\
    \addlinespace
    Decision   & override requires a single \textsc{Fit} & yes \\
               & override min.\ members                  & 5 \\
               & tie-break                               & confidence, then member count \\
    \addlinespace
    Runtime    & workers                                 & 4 \\
    \bottomrule
  \end{tabular}
\end{table}

\subsection{Prompt Templates}
\label{supp:prompts}
PaperRouter-Agent makes two LLM calls: a per-folder \emph{Inspector} verification and,
when no folder is a content-grounded \textsc{Fit}, a single-shot \emph{baseline}
classifier. The prompts are paper-framed. On LaMP-2 the paper is a movie and the
folder is a tag.

The \emph{Inspector} prompt:
\begin{lstlisting}[style=prompt]
You are deciding whether a new paper belongs in one specific folder of a researcher's personal library.

You are given the new paper, the folder's NAME, and a sample of the papers ALREADY in that folder
(this sample MAY BE EMPTY if the folder has no papers yet).

Judge membership PRIMARILY from what the folder ACTUALLY CONTAINS -- its member papers -- because the
name alone may be a shorthand, a deep-hierarchy node, or otherwise misleading. The same folder name can
mean different things to different researchers; infer this researcher's sense of it from THEIR members.
When the folder has few or no members to learn from, fall back to the folder's NAME and judge by the
topic it denotes.

New paper:
  Title: {paper_title}
  Abstract: {paper_abstract}

Candidate folder: "{folder_name}"
Sample of its current member papers:
{member_list}

Choose exactly one label:
- FIT      -- the new paper clearly belongs here: with these members, or -- if there are none -- with the
              topic the folder name denotes.
- NO_FIT   -- the new paper does not belong here.
- ABSTAIN  -- the members EXIST but are too topically heterogeneous to establish what this folder is
              about (e.g. grouped by venue, reading status, or source rather than by subject), AND the
              name gives no usable topic either.

Prefer the members' content whenever it is informative. Do NOT abstain merely because the folder has
few or no members -- in that case judge from the name. Reserve ABSTAIN for genuinely incoherent folders.

Return JSON only, no prose:
{"label": "FIT" | "NO_FIT" | "ABSTAIN", "confidence": <0.0-1.0>, "rationale": "<one sentence; cite the member papers' content when available, otherwise the folder name>"}
\end{lstlisting}

The single-shot \emph{baseline} prompt:
\begin{lstlisting}[style=prompt]
You classify a movie into exactly ONE tag, by its plot/content alone.
Choose the single best-fitting tag from this list:
{tag_list}

Movie:
{plot}

Return JSON only, no prose: {"tag": "<exactly one tag copied verbatim from the list above>"}
\end{lstlisting}

\end{document}